\pdfoutput=1

\documentclass[11pt]{article}

\usepackage{acl}

\usepackage{times}
\usepackage{latexsym}
\usepackage{multirow}
\usepackage{tabularx}
\usepackage{booktabs}
\usepackage{graphicx}
\usepackage{amsmath}

\usepackage[T1]{fontenc}

\usepackage[utf8]{inputenc}

\usepackage{microtype}

\usepackage{fancyhdr}
\usepackage{lipsum} 

\fancypagestyle{firstpage}{
    \fancyhf{} 
    \chead{\textcolor{gray}{\textit{This paper has been accepted for publication at the ACL Student Research Workshop 2024}}} 
}

\title{Curriculum Learning for Small Code Language Models}

\author{
Marwa Naïr$^{1,2,*}$,  {\bf Kamel Yamani$^{1,2,*}$},  {\bf Lynda Said Lhadj$^{1}$},  {\bf Riyadh Baghdadi$^{2}$}   \\ 
$^1$Ecole nationale Supérieure d’Informatique (ESI), Algeria \\
$^2$New York University Abu Dhabi (NYUAD), United Arab Emirates \\ 
\texttt{\{jm\_nair, jm\_yamani, l\_said\_lhadj\}@esi.dz} \\
\texttt{baghdadi@nyu.edu}
}

\begin{document}
\thispagestyle{firstpage}
\maketitle

\def\thefootnote{*}\footnotetext{Both authors contributed equally to this work and share first authorship.}\def\thefootnote{\arabic{footnote}}

\begin{abstract}

Code language models have emerged as useful tools for various programming tasks, yet they often struggle when it comes to complex ones.
In this paper, we explore the potential of curriculum learning in enhancing the performance of these models.
While prior research has suggested that curriculum learning does not necessarily help in improving the performance of language models, our results surprisingly show that this may not be the case for code language models.  We demonstrate that a well-designed curriculum learning approach significantly improves the accuracy of small decoder-only code language models on the task of code execution, while its effect on code completion is less significant. 
To explore the potential of curriculum learning, we train multiple GPT models with 1 million parameters each to predict the next token and evaluate them on code completion and execution tasks.
Our contributions include proposing a novel code difficulty assessment metric by combining software code measures, investigating the effectiveness of Curriculum Learning for code language models, and introducing a Novel Curriculum Learning schedule that enhances the performance of small decoder-only language models in code execution tasks. The results of this paper open the door for more research on the use of curriculum learning for code language models. 

\end{abstract}

\section{Introduction}
\label{sec:introduction}
With the advent of large language models (LLMs) like GPT-3
\citep{brown_language_2020}, auto-regressive decoder-only architectures have become dominant in language modeling. These models have shown significant improvement over state-of-the-art performance on a wide range of natural language tasks. Accordingly, previous work \cite{chen_evaluating_2021, lu_codexglue_2021, nijkamp_codegen_2022, zheng_codegeex_2023} has introduced such architectures for code modeling, motivated by the software naturalness hypothesis \citep{hindle_naturalness_2016, buratti_exploring_2020}, which suggests that programming languages can be understood and generated like natural languages \citep{xu_survey_2022}.

However, these models often struggle with complex tasks such as understanding code and reasoning about it, which remains a challenge for them. \citet{austin_program_2021} evaluated the ability of large language models to predict the output of ground-truth programs. The authors found that the few-shot execution performance of their largest model, with 137 billion parameters, never exceeded 29\% accuracy across various prompt configurations. Fine-tuning on an code execution dataset resulted in only modest improvements, with the best configuration achieving 28.2\% accuracy. 

In this context, we investigate whether Curriculum Learning (CL) - training models on simpler examples first before gradually increasing difficulty - can improve the performance of decoder-only language models' trained on source code. We assume that training language models using CL will lead to better performance compared to conventional training. We focus on small-scale models, which allows us to experiment with different setups and iterate quickly.

Prior research has investigated the use of curriculum learning for language model pre-training, finding no substantial evidence to support its effectiveness \cite{campos_curriculum_2021}. However, the potential benefits of this approach in the context of Code Intelligence
\citep{xu_survey_2022}, remain relatively unexplored. In contrast to these earlier findings, our investigation indicates that the advantages of CL may be more task-dependent. Particularly, we show that while CL exhibits potential in enhancing code execution capabilities, its influence on code completion tasks is less significant.

More specifically, we follow an incremental study where we generate a Python code dataset, design a code difficulty assessment metric which enables us to categorize our dataset into three levels: “easy”, “medium”, and “hard”. Based on these levels, we propose three-stage Curriculum Learning (CL) schedules to train our code language models. 

To further illustrate the challenges posed by complex code examples, we evaluated the Code Llama 13B model \citep{roziere_code_2024} on our "hard level" test set, where it achieved only 39.06\% accuracy. This evaluation highlights the limitations of current LLM-based code modeling approaches, which still struggle to effectively capture the semantics of source code.

Our results indicate that performance on code execution can indeed be improved when we design a good curriculum schedule and use a robust code difficulty metric. However, when it comes to code completion tasks, the impact of CL is less pronounced. This suggests that the benefits of CL may not be present across all tasks, but rather, depend on the specific nature of the task.

Believing this can advance the research on code language models, we have open-sourced our datasets\footnote{\url{https://tinyurl.com/TinyPyD}}, models and  source code\footnote{\url{https://tinyurl.com/CL4SCLM}}.

Our main contributions can be summarized as follows:
\begin{itemize}
    \item First, we propose a code difficulty assessment metric combining software code measures. 
    \item Second, we explore the potential of Curriculum Learning for auto-regressive code language models by investigating numerous curriculum schedules. 
    \item Finally, we propose a Novel Curriculum Learning schedule that improves small decoder-only language models' performance on code execution. 
\end{itemize}

\section{Overview}
\label{sec:overview}

\begin{figure*}
    \centering
    \includegraphics[scale=0.78]{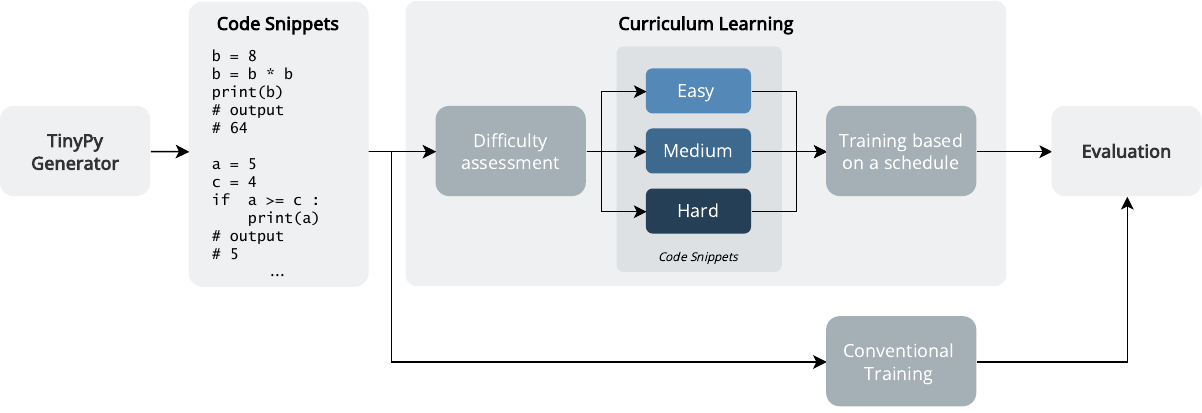}
    \caption{Overview of Our Approach : We begin by generating code snippets using TinyPy Generator. Next, we assess the difficulty of the generated code snippets using the Overall Metric we propose and categorize the data into three levels of difficulty: easy, medium, and hard. Our 1M parameters decoder-only language models are trained following various Curriculum Learning schedules. We then compare their performance to a 1M baseline model trained on all the data simultaneously, with all three levels shuffled.}
    \label{fig:overview}
\end{figure*}

In order to explore whether using Curriculum Learning can improve the performance of decoder-only language models trained on code, we adopt the following methodology (Presented in \autoref{fig:overview}) : We first generate data (consisting of code snippets followed by their outputs) focusing on a subset of the Python programming language, which allows us to reduce the vocabulary size (\autoref{sec:dataset}).  We then assess the difficulty of the generated code snippets using our proposed code difficulty metric, which we refer to as the Overall Metric (OM) (\autoref{sec:code}) and split the data into three levels - easy, medium, and hard. Next, the models are trained on different Curriculum Learning schedules (\autoref{sec:curriculum}). Finally, we evaluate the performance of the models based on  token-level and line-level code completion as well as code execution, and compare them to a baseline model trained on all levels of data shuffled together (\autoref{sec:experiments}).

Additionally, to investigate the effect of Curriculum Learning on larger pretrained models, we finetuned Code Llama 7B \citep{roziere_code_2024} using our best Curriculum Learning schedule and compared it with a baseline finetuning approach where all levels of data are shuffled together (\autoref{sec:experiments}).

\section{Code Difficulty Metric}
\label{sec:code}

Determining the difficulty of code is not straightforward. It requires a quantitative measure, which can be provided by commonly used software engineering metrics like Cyclomatic Complexity (CC) and Halstead Difficulty (HD). CC, proposed by \citet{mccabe_complexity_1976}, quantifies the number of linearly independent paths through a program’s source code. On the other hand, HD, introduced by \citet{halstead_elements_1977}, is calculated using the number of operators and operands present in the code. These established metrics allow for the numerical evaluation of code difficulty. However, their independent use may not fully capture the overall difficulty of the code.

Therefore, we have designed a new metric, referred to as the Overall Metric (OM), which is the average of CC and HD (see \autoref{eq:OM}). The idea behind creating OM is to have a more comprehensive measure of difficulty that takes into account both structural complexity via CC and operational complexity via HD.

\begin{equation}
\label{eq:OM}
OM = \frac{CC + HD}{2}
\end{equation}

\section{Dataset Generation Process}
\label{sec:dataset}

\subsection{Automatic Python Code Generation}

To generate the data for training code language models in a curriculum learning setting, we used \textbf{TinyPy Generator} \citep{yamani_automatic_2024}, an automatic Python code generation tool developed by us. This tool uses context-free grammars to generate synthetic syntactically correct Python programs, focusing on a constrained subset of Python that includes assignments, conditionals, loops, and print statements. This vocabulary constraint decreases the Embeddings dimension, leaving more capacity for Transformer blocks while maintaining a small number of parameters, as pre-training loss decreases insignificantly without Transformer blocks \cite{deshpande_honey_2023}.

 TinyPy Generator not only generates code snippets but also executes and writes them along with their respective outputs (expressed in comments) to a file. By training the model on code followed by its output, we assume that this helps the model to better get the connection between the code and its intended function.

\subsection{Analysis of Generated Code Snippets}
\label{sec:analysis}

\begin{figure}
    \centering
    \includegraphics[scale=0.54]{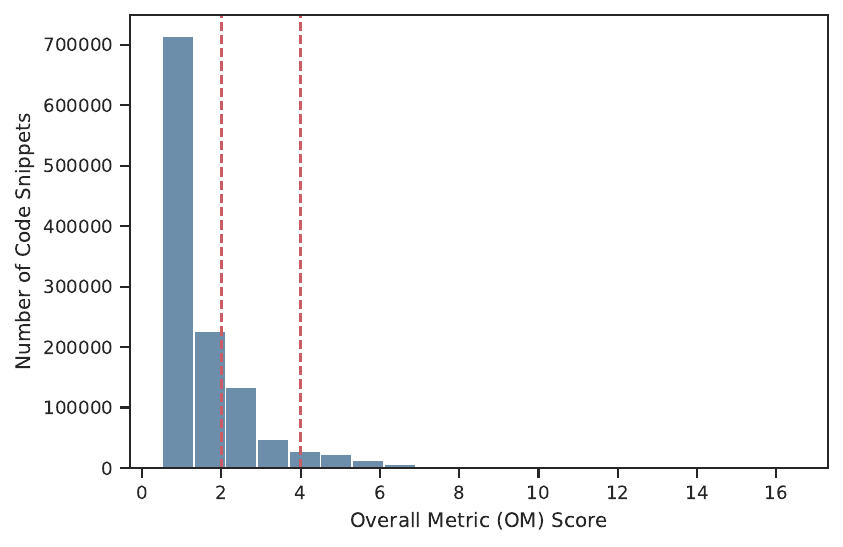}
     \caption{Distribution of Overall Metric (OM) Scores for the Initial Set of Generated Snippets.}
    \label{fig:dist}
\end{figure}

We first used TinyPy Generator to generate 1,200,000 random code snippets (examples shown in Figure~\ref{fig:snippets}). We then categorized these automatically generated programs based on their difficulty according to the OM metric. More precisely, we had to determine optimal thresholds to divide the generated snippets into three levels of difficulty: easy, medium, and hard. The Visualisation of the distribution of OM scores for the generated snippets (depicted in \autoref{fig:dist}) revealed that most fell into the easy category with $OM < 2$. A smaller subset were of medium difficulty with $2 \leq OM < 4$, and the smallest group were hard snippets with $OM \geq 4$. This analysis helped us understand the OM score ranges for the code snippets produced by TinyPy Generator. Additionally, it allowed us to determine the thresholds for easy, medium, and hard snippets.

\subsection{Dataset Creation}
\label{sec:data-creation}
\begin{figure}[h]
    \centering
    \includegraphics{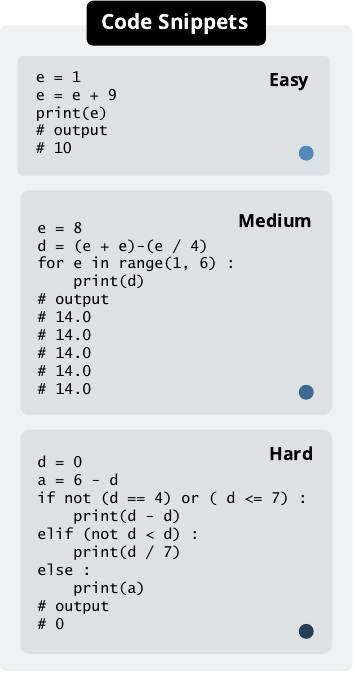}
     \caption{One code snippet example from each difficulty level (the examples are chosen arbitrarily). More examples are presented in \autoref{sec:examples}.}
    \label{fig:snippets}
\end{figure}

Building on the insights from the analysis, we proceeded to create a balanced dataset for the training of our language models. More precisely, we produced a total of 400k snippets for each level, culminating in a dataset of 1.2M snippets in total, as shown in \autoref{fig:snippets} (more examples are presented in \autoref{sec:examples}). Then, each level’s dataset was randomly partitioned into training, validation, and testing sets. After that, we  proceeded to create the ‘ALL levels’ dataset, which is a shuffled concatenation of all train, test, and validation sets from each level into the train, test, and validation sets of the ALL dataset. Additional details about the final datasets are provided in \autoref{tab:data}.

\begin{table}[h]
\centering
\small
\begin{tabularx}{\columnwidth}{*{5}{>{\centering\arraybackslash}X}}
\toprule
Difficulty Level & \textbf{Easy} & \textbf{Medium} & \textbf{Hard} & \textbf{ALL levels} \\ 
\midrule

\multirow{2}{*}{\shortstack{Train \# \\ \scriptsize{in tokens}}} & 340,000 & 340,000 & 340,000 & 1,020,000 \\
& \scriptsize{22,438,558} & \scriptsize{30,777,288} & \scriptsize{42,719,202} & \scriptsize{95,935,048} \\\\

\multirow{2}{*}{\shortstack{Val \# \\ \scriptsize{in tokens}}} & 52,000 & 52,000 & 52,000 & 156,000 \\
& \scriptsize{3,436,602} & \scriptsize{4,710,195} & \scriptsize{6,533,573} & \scriptsize{14,680,370} \\\\

\multirow{2}{*}{\shortstack{Test \# \\ \scriptsize{in tokens}}} & 8,000 & 8,000 & 8,000 & 24,000 \\
& \scriptsize{527,649} & \scriptsize{724,530} & \scriptsize{1,005,030} & \scriptsize{2,257,209} \\
\bottomrule
\end{tabularx}
\caption{Training data statistics : The number of code snippets (training, validation, and test) and their corresponding token counts for each difficulty level - Easy, Medium, Hard, and cumulatively for All Levels.}
\label{tab:data}
\end{table}

\section{Curriculum Learning Schedules}
\label{sec:curriculum}
\begin{figure}[h]
    \centering
    \includegraphics[scale=0.9]{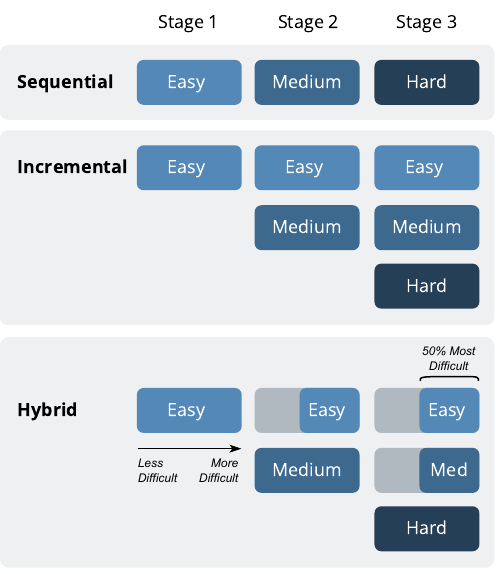}
     \caption{Our three curriculum learning schedules. Sequential progresses from easy to hard snippets sequentially. Incremental starts with easy snippets, gradually adding harder ones. The Hybrid schedule starts with easy snippets, then adds a mix of the hardest easy snippets and medium snippets, and finally combines the hardest snippets from the easy and medium levels with hard snippets.
    }
    \label{fig:schedules}
\end{figure}

Curriculum learning (CL) is a training strategy that presents easier or simpler examples earlier in training and gradually increases the difficulty of examples over time.
This section details the Curriculum Learning schedules we propose, namely: Sequential, Incremental, and Hybrid, illustrated in \autoref{fig:schedules}. Each schedule is divided into three stages. After completing a stage, we reset the learning rate and optimizer before continuing training on the data for the next stage. The three schedules are defined as follows:

\subsection{Sequential curriculum learning schedule}

In the Sequential Curriculum Learning schedule, the model is initially trained on the 'easy' level data for a fixed number of iterations. After this stage, the model moves to the ‘medium’ level data. After another fixed number of iterations, the model finally transitions to the ‘hard’ level.

\subsection{Incremental curriculum learning schedule}

The Incremental Curriculum Learning schedule progressively introduces more complex data into the training set. The model starts with the ‘easy’ level data for a fixed number of iterations. Once this stage is complete, the ‘medium’ level data is added to the training set for another fixed number of iterations. Upon completion of this stage, ‘hard’ level data is incorporated.

\subsection{Hybrid curriculum learning schedule}

The Hybrid Curriculum Learning schedule is a blend of the Sequential and Incremental schedules. In the first stage, the model is trained exclusively on the ‘easy’ level data for a certain number of iterations. In the second stage, a combination of the top 50\% most difficult examples from the ‘easy’ level data and the ‘medium’ level data is used for training. In the final stage, we combine both top 50\% most difficult examples from the ‘easy’ and ‘medium’ levels with the ‘hard’ level data.

\section{Experimental Setup}
\label{sec:setup}
In this section, we describe our experimental setup for evaluating the effectiveness of curriculum learning for code language models, including the model architecture we used, our training process, and the evaluation tasks and metrics we employed.

\subsection{Model Architecture}
For our models, we employ NanoGPT\footnote{The original NanoGPT \cite{karpathy_karpathynanogpt_2022} is licensed under the permissive MIT License, allowing modification and distribution.} \cite{karpathy_karpathynanogpt_2022}, a small version of the GPT model family. The primary reason for this choice is its ability to train from random initialization (from scratch) under a variety of settings, allowing for rapid iteration. NanoGPT employs a decoder-only transformer architecture, comprising six self-attention layers, six heads, and an embedding dimension of 384. This results in approximately 10.6 million parameters. We modify the model by reducing the embedding dimension to 120 and setting the block size to 256, which results in a model with around 1 million parameters. The model uses a vocabulary size of 41 and does not include bias in its linear layers. We employ character-level tokenization and absolute position encoding. 

\subsection{Training Details}

All our models are trained from scratch using the conventional next-token prediction objective.  The hyperparameters for each model were selected based on minimizing the validation loss. \\

\textbf{Baseline Model}:  The baseline model is trained on all the data simultaneously, with all three levels shuffled. Given the small size of our model, we do not find it necessary to employ dropout for regularization. The batch size is set to 64, the learning rate is set to 1e-3, and the AdamW optimizer is used for training. The learning rate decay is implemented using milestones set at 70\%, 80\%, and 90\% of the total number of iterations, which is 120k.\\

 \textbf{Models Trained with Curriculum Learning (CL)}: These models also do not use dropout and have a batch size of 64. However, the number of iterations varies for each stage, with the total summing up to 120k iterations (See \autoref{tab:train}). Note that we tested various iterations settings and reported the best. For each stage, the learning rate is set to 1e-3, and is decayed using the same milestone percentages as the baseline model. The AdamW optimizer is used for training.

\begin{table}[h]
\centering
\small
\begin{tabularx}{\columnwidth}{>{\raggedright\arraybackslash}p{0.35\columnwidth}  
                                 >{\centering\arraybackslash}p{0.3\columnwidth}
                                 >{\centering\arraybackslash}X}
\toprule
\textbf{Model}  & \textbf{Iterations per stage} & \textbf{Total Iterations} \\
\midrule
Baseline  & - & 120k\\
Sequential CL  & 40k, 40k, 40k & 120k\\
Incremental CL  & 25k, 30k, 65k & 120k\\
Hybrid CL  & 20k, 30k, 70k & 120k\\
\bottomrule
\end{tabularx}
\caption{Details of Training Iterations for our models. ‘Iterations per stage’ denotes the number of iterations for each stage for models trained using CL.}
\label{tab:train}
\end{table}

\subsection{Evaluation tasks and metrics}

\begin{figure}[h]
    \centering
    \includegraphics[scale=1]{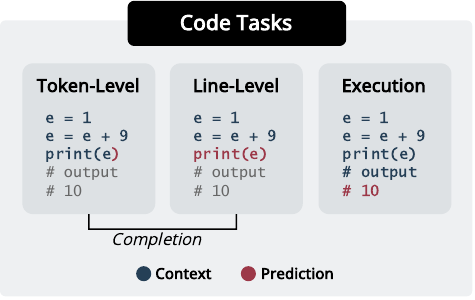}
     \caption{Experimental Evaluation on Three Code Tasks: Token-Level Completion, Line-Level Completion, and Code Execution}
    \label{fig:tasks}
\end{figure}

To evaluate the effectiveness of Curriculum Learning for improving code language models, we assess their performance on three key tasks: token-level completion, line-level completion, and code execution, as presented in \autoref{fig:tasks}.

\subsubsection{Code Completion}
We evaluate code completion performance at two levels, inspired by the approach used in CodeXGLUE \cite{lu_codexglue_2021}:

\begin{itemize}
\item \textbf{Token-level}: Similar to \cite{lu_codexglue_2021}, models are evaluated on completing the next token in the incomplete snippet.
\item \textbf{Line-level}: We slightly modified the line-level task from \cite{lu_codexglue_2021}. We provide the model with the previous lines of the incomplete snippet and let it generate the next line.
\end{itemize}

For both levels, we report the \textbf{Accuracy}, which measures whether the model's output exactly matches the expected output. For the Line completion task, we also report the \textbf{Edit Similarity (ES)} calculated using Levenshtein distance between the model's predicted line and the expected line.

\subsubsection{Code Execution}
To assess the code execution abilities of our models, we utilize the 'ALL levels' test set, as detailed in \autoref{sec:data-creation}. The models are prompted with the code portion of the test snippets, stopping at the ‘\# output’ comment to exclude the output and let the model predict it. The model generates the output one token at a time, as described in \autoref{sec:generation}. We employ the Output Accuracy as our evaluation metric, which checks if the generated output exactly matches with the expected output from executing the code. The accuracy is calculated for each difficulty level and an overall accuracy is computed across all levels.

\section{Experiments and Results}
\label{sec:experiments}

\subsection{Correlating OM with Model Learning Capabilities}

To validate the effectiveness of OM in assessing the difficulty of code snippets, we generated six conceptual levels of complexity, based on programming concepts:
(1) assignments with simple arithmetic; (2) assignments with advanced arithmetic expressions; (3) simple if-elif-else statements; (4) advanced if-elif-else statements with arithmetic expressions; (5) simple for loops; and (6) advanced for loops with arithmetic expressions.

We trained and evaluated models with less than 1 million parameters on each level and reported their average accuracy in \autoref{tab:OM}. The results showed an inverse relationship between OM and accuracy, confirming OM’s effectiveness in ranking code snippet difficulty.

\begin{table}
\centering
\small
\begin{tabular}{ccc}
\hline
\textbf{Level} & \textbf{OM} & \textbf{Average Accuracy} \\
\hline
1 & 0.85 & 96.65\% \\
2 & 0.98 & 87.33\% \\
3 & 1.77 & 58.43\% \\
4 & 3.5 & 50.29\% \\
5 & 1.0 & 83.73\% \\
6 & 1.38 & 73.19\% \\
\hline
\end{tabular}
\caption{Average Overall Metric (OM) Score vs. Average Accuracy Achieved by models under 1M parameters at Each level from 1 to 6.}
\label{tab:OM}
\end{table}

\subsection{Code Completion} 
To test the effectiveness of CL for code completion, we compared models trained with CL with our baseline. As shown in  \autoref{tab:completion}, the incremental approach leads to a minor gain in token-level accuracy over the baseline. Similarly, the hybrid curriculum achieves small improvements in Line-level accuracy of 0.3\% and edit similarity of 0.5. While these results demonstrate that curriculum learning can provide some benefits, the improvements are not significant enough to conclusively state its effectiveness for code completion. 
\begin{table}[h]
\centering
\small
\begin{tabularx}{\columnwidth}{>{\raggedright\arraybackslash}p{0.26\columnwidth}
                                >{\centering\arraybackslash}p{0.21\columnwidth}
                                >{\centering\arraybackslash}p{0.15\columnwidth}
                                >{\centering\arraybackslash}p{0.15\columnwidth}}
\toprule
Model & Token-Level  & \multicolumn{2}{c}{Line-Level} \\
 & Accuracy & Accuracy & ES \\
\midrule
Baseline & 81.23\%  & 41.74\% & 74.15 \\
Sequential CL & 75.64\%  & 25.84\% & 66.96 \\
Incremental CL & \textbf{81.27\%} & 42.01\% & 74.25 \\
Hybrid CL  & 81.13\%  & \textbf{42.04\%} & \textbf{74.65} \\
\bottomrule
\end{tabularx}
\caption{Performance Evaluation of Our Models on Code Completion Tasks, measured in terms of Token-Level Accuracy and  Line-Level Accuracy and Edit Similarity (ES).}
\label{tab:completion}
\end{table}

\subsection{Code Execution} 

\subsubsection{Performance on All Levels} 

To determine the impact of different curriculum learning (CL) strategies on code execution performance, we compared models trained with incremental, sequential, and hybrid CL schedules against a baseline model trained on all difficulty levels simultaneously. As shown in Table \ref{tab:accuracy}, the hybrid CL approach achieves the best performance, with significant gains over the baseline on medium and hard test sets. The incremental CL model also improves upon the baseline overall. However, sequential CL enables some learning of advanced concepts but reduces overall performance. In conclusion, our results demonstrate that a well-designed curriculum, especially the hybrid schedule, substantially outperforms conventional training without CL for code execution tasks.

\begin{table*}[h] 
\centering 
\begin{tabular}{ccccc}
\hline
\textbf{Model} & \textbf{ALL} & \textbf{Easy} & \textbf{Medium} & \textbf{Hard}  \\
\hline
Baseline & 74.58\% & 80.44\% & 76.09\%  & 67.22\% \\
Sequential CL & 62.56\% & 46.47\% & 70.73\% & 70.47\% \\
Incremental CL & 76.79\% & 82.63\% & 77.68\% & 70.06\% \\
Hybrid CL & \textbf{79.23\%} &\textbf{ 82.84\%} & \textbf{80.79\%} & \textbf{74.04\%} \\
\hline 
\end{tabular}
\caption{Output Accuracy of Our Models - Baseline, Sequential CL, Incremental CL, and Hybrid CL - on different levels of difficulty (Easy, Medium, Hard) and their overall accuracy (ALL).} 
\label{tab:accuracy}
\end{table*}

\subsubsection{Performance on the "hard" Level} 

To evaluate the ability of curriculum learning (CL) to prepare a model for complex tasks, we compared the performance of models trained with CL to a model trained exclusively on the “hard” training set for 120k iterations. The comparison is conducted on the "hard" test set, which contains the most complex examples in our dataset. The results are presented in \autoref{tab:complex}. We notice that all three CL approaches substantially outperformed the model trained exclusively on hard data, with the hybrid CL method achieving the highest accuracy of 74.04\%.This shows that CL is more effective than conventional hard-only training for preparing models to perform well on complex code execution examples.

\begin{table}[h] 
\centering
\begin{tabular}{cc}
\hline
\textbf{Model} & \textbf{Accuracy}  \\
\hline
Trained on hard level only & 61.78\%  \\
Sequential CL & 70.47\%  \\
Incremental CL & 70.06\%   \\
Hybrid CL & \textbf{74.04\%}  \\
\hline 
\end{tabular}
\caption{Output Accuracy of a model trained exclusively on hard level versus models trained using CL schedules: ‘Sequential’, ‘Incremental’, and ‘Hybrid’,  when tested on hard examples.} 
\label{tab:complex}
\end{table}

\subsection{Investigating the Effect of Curriculum Learning on Larger Pretrained Models}
To further validate the effectiveness of curriculum learning (CL) observed in our earlier experiments, we extended our evaluation by fine-tuning the Code Llama 7B model \citep{roziere_code_2024}. We compared the performance of a model fine-tuned on the 'ALL' dataset (referred to as 'CodeLlama Baseline') with a model fine-tuned using the hybrid CL technique (referred to as 'CodeLlama CL'). The results consistently reflected the improvements noted in smaller models.

For code completion tasks, as shown in \autoref{tab:completion-finetuning}, the CodeLlama CL model demonstrated minor improvements over the baseline model. For code execution, as illustrated in \autoref{tab:execution-finetuning}, the CodeLlama CL model significantly outperformed the baseline model.

These findings validate that CL advantages scale to larger pretrained models. The consistent gains across model sizes highlight our CL approch's generalizability for enhancing code understanding in auto-regressive language models.

\begin{table}[h]
\centering
\small
\begin{tabularx}{\columnwidth}{>{\raggedright\arraybackslash}p{0.26\columnwidth}
>{\centering\arraybackslash}p{0.21\columnwidth}
>{\centering\arraybackslash}p{0.15\columnwidth}
>{\centering\arraybackslash}p{0.15\columnwidth}}
\toprule
Model & Token-Level  & \multicolumn{2}{c}{Line-Level} \\
 & Accuracy & Accuracy & ES \\
\midrule
CodeLlama Baseline & 72.00\% & 32.83\% & 70.11 \\
CodeLlama CL & \textbf{72.73\%} & \textbf{33.54\%} & \textbf{70.84} \\
\bottomrule
\end{tabularx}
\caption{Fine-tuning results of Code Llama 7B with and without hybrid curriculum learning (CL) for code completion tasks.}
\label{tab:completion-finetuning}
\end{table}

\begin{table}[h] 
\centering
\begin{tabular}{cc}
\hline
\textbf{Model} & \textbf{Accuracy}  \\
\hline
CodeLlama Baseline & 81.29\% \\
CodeLlama CL & \textbf{85.18\%} \\
\hline 
\end{tabular}
\caption{Fine-tuning results of Code Llama 7B with and without hybrid curriculum learning (CL) for code execution.}
\label{tab:execution-finetuning}
\end{table}

\section{Discussion}
\label{sec:discussion}

We designed a code difficulty metric combining software measures, referred to as OM,  to categorize generated programs into easy, medium and hard levels. The inverse correlation between the OM scores and the model accuracies validates its effectiveness for program difficulty assessment. An interesting observation is that conditionals posed more difficulty for models than loops, contrary to expectations. This suggests certain language features are inherently harder to learn for models.

This categorization allowed us to explore various three-stage curriculum schedules for model training. Our experiments revealed that the hybrid technique achieves much higher output accuracy compared to the conventional training baseline, especially on complex code, indicating its effectiveness in incrementally developing model capabilities. However, the sequential strategy, while helping models learn hard concepts, suffers a loss in overall accuracy. This highlights the importance of curriculum design : simply progressing from easy to hard tasks does not guarantee gains.

In the context of code completion tasks, the influence of CL is not as significant as expected. This implies that the advantages of CL may not be applicable to all tasks, but instead, they may vary based on the particular characteristics of the task.

Furthermore, our fine-tuning experiments with the Code Llama 7B model further validated the effectiveness of curriculum learning. While the gains in code completion tasks were minor, the hybrid CL approach significantly improved code execution performance. These findings reinforce our findings that a well-designed curriculum can enhance model capabilities, especially for complex tasks, even when scaling to larger models.

\section{Related Works}
\label{sec:relatedworks}
\subsection{Code Language Models}

The application of pre-trained Transformers in code processing can be traced back to dates before decoder-only auto-regressive models became dominant. These models have consistently delivered state-of-the-art results across a wide range of tasks, including code summarization, generation, and translation \cite{xu_survey_2022}. Such examples include encoders like CuBERT \cite{kanade_learning_2020}, CodeBert \cite{feng_codebert_2020} and  GraphCodeBERT \cite{guo_graphcodebert_2020}. The use of the encoder-decoder architecture have also been proposed with models like : CodeT5  \cite{wang_codet5_2021}, CodeT5+ \cite{wang_codet5_2023} and AlphaCode \cite{li_competition-level_2022}.

Following the introduction of GPT-3 \cite{brown_language_2020}, autoregressive decoder-only language models have taken a leading role in the field of language modeling. Consequently, a multitude of studies have been published proposing the use of such architectures for code. Codex by OpenAI \cite{chen_evaluating_2021}, one of the largest language models for code, is trained on public repositories on Github across multiple programming languages. Other notable attempts include CodeGPT \cite{lu_codexglue_2021}, CodeGen \cite{nijkamp_codegen_2022}, PolyCoder \cite{xu_systematic_2022}, CodeGeeX \cite{zheng_codegeex_2023}, and Code Llama \cite{roziere_code_2024}.

\subsection{Curriculum Learning}

Prior work has investigated curriculum learning \cite{elman_learning_1993, sanger_neural_1994, bengio_curriculum_2009} for the pre-training of language models. The paper introduced by \citet{li_stability-efficiency_2022} discusses the concept of Sequence Length Warmup, a method that uses CL for stable training of GPT models with larger batches and learning rates. This significantly reduces data and time requirements for pre-training. Additionally, the effectiveness of curriculum learning for pre-training BERT models has been explored in several studies \cite{press_shortformer_2021, zhang_reducing_2021, campos_curriculum_2021, nagatsuka_pre-training_2021, nagatsuka_length-based_2023}. The results have been mixed. Some research shows curriculum learning can accelerate convergence, shorten training time, and boost accuracy while other studies do not find these advantages. 

\section{Conclusion}
\label{sec:conlusions}

In this paper, we explored the potential of curriculum learning in enhancing the performance of code language models, given their struggle with complex tasks. 

First, we generated a dataset of Python code using the TinyPy Generator. Second, we designed a code difficulty metric (OM) combining software complexity measures, and validated its efficacy in assessing program difficulty. Third, we used the OM to categorize programs into easy, medium, and hard levels and explored various curriculum schedules. Finally, we evaluated our models on code completion and execution tasks and compared them to a baseline trained on all the data shuffled. Our results show that certain curriculum learning strategies can significantly improve language models' performance on code execution, compared to conventional training. Nonetheless, for code completion, the gains from CL were not as significant as expected. 

Additionally, our fine-tuning experiments with the Code Llama 7B model reinforced these findings, demonstrating that CL can lead to significant improvements in code execution tasks even for larger models.

In conclusion, our investigation shows that thoughtfully implemented curriculum learning can improve generative code language models' performance on code execution tasks. Yet, its impact is less noticeable in code completion tasks. This suggests that curriculum learning's effectiveness may vary depending on the task's specific characteristics. Overall, our work highlights the potential of curriculum learning to enhance language models for complex code reasoning.

\section*{Limitations}
\label{sec:limitations}

Some limitations provide avenues for future work. Our study was restricted to a subset of Python. Testing curriculum techniques on all the Python language could reveal if its advantages generalize across the entire language. Additionally, our focus solely on Python code represents another limitation. Exploring whether curriculum learning improves performance for other programming languages merits investigation.

Nevertheless, within the defined scope, our findings strongly suggest curriculum learning is a promising training paradigm for boosting code execution performance. The hybrid curriculum schedule we propose offer a sound starting point for integrating curriculum learning into code language model development. Extending this approach by addressing the above limitations provides rich opportunities for future work.

\section*{Ethical Statement}
This work was carried out in compliance with ethical standards. The TinyPy dataset used for training and evaluation were automatically generated using context free grammars, rather than scraping potentially sensitive or copyrighted data from public code repositories. As a result, the data does not raise privacy, copyright infringement, or dual use concerns. Additionally, there was no human annotation of the data, so no crowdsourcing that would require ethical considerations around recruitment, compensation, or informed consent.

We have also tried to minimize environmental costs like high energy usage, carbon emissions, and electronic waste from GPUs by focusing experiments on small models that require far less computation. All our experiments were conducted on an environmentally-friendly cluster.

One key risk is that of malicious use, where bad actors could leverage powerful code generation systems to automatically produce harmful software like viruses or bots. Even without harmful intent from researchers, releasing and open-sourcing our curriculum learning methodology and model code could enable this misuse if proper safeguards are not implemented.

\section*{Acknowledgements}
This work was supported in part through the NYU IT High Performance Computing resources, services, and staff expertise.

\bibliographystyle{acl_natbib}
\bibliography{references}

\begin{thebibliography}{31}
\expandafter\ifx\csname natexlab\endcsname\relax\def\natexlab#1{#1}\fi

\bibitem[{Austin et~al.(2021)Austin, Odena, Nye, Bosma, Michalewski, Dohan, Jiang, Cai, Terry, Le, and Sutton}]{austin_program_2021}
Jacob Austin, Augustus Odena, Maxwell Nye, Maarten Bosma, Henryk Michalewski, David Dohan, Ellen Jiang, Carrie Cai, Michael Terry, Quoc Le, and Charles Sutton. 2021.
\newblock \href {https://doi.org/10.48550/arXiv.2108.07732} {Program {Synthesis} with {Large} {Language} {Models}}.
\newblock ArXiv:2108.07732 [cs].

\bibitem[{Bengio et~al.(2009)Bengio, Louradour, Collobert, and Weston}]{bengio_curriculum_2009}
Yoshua Bengio, Jérôme Louradour, Ronan Collobert, and Jason Weston. 2009.
\newblock \href {https://doi.org/10.1145/1553374.1553380} {Curriculum learning}.
\newblock In \emph{Proceedings of the 26th {Annual} {International} {Conference} on {Machine} {Learning}}, {ICML} '09, pages 41--48, New York, NY, USA. Association for Computing Machinery.

\bibitem[{Brown et~al.(2020)Brown, Mann, Ryder, Subbiah, Kaplan, Dhariwal, Neelakantan, Shyam, Sastry, Askell, Agarwal, Herbert-Voss, Krueger, Henighan, Child, Ramesh, Ziegler, Wu, Winter, Hesse, Chen, Sigler, Litwin, Gray, Chess, Clark, Berner, McCandlish, Radford, Sutskever, and Amodei}]{brown_language_2020}
Tom Brown, Benjamin Mann, Nick Ryder, Melanie Subbiah, Jared~D Kaplan, Prafulla Dhariwal, Arvind Neelakantan, Pranav Shyam, Girish Sastry, Amanda Askell, Sandhini Agarwal, Ariel Herbert-Voss, Gretchen Krueger, Tom Henighan, Rewon Child, Aditya Ramesh, Daniel Ziegler, Jeffrey Wu, Clemens Winter, Chris Hesse, Mark Chen, Eric Sigler, Mateusz Litwin, Scott Gray, Benjamin Chess, Jack Clark, Christopher Berner, Sam McCandlish, Alec Radford, Ilya Sutskever, and Dario Amodei. 2020.
\newblock \href {https://papers.nips.cc/paper/2020/hash/1457c0d6bfcb4967418bfb8ac142f64a-Abstract.html} {Language {Models} are {Few}-{Shot} {Learners}}.
\newblock In \emph{Advances in {Neural} {Information} {Processing} {Systems}}, volume~33, pages 1877--1901. Curran Associates, Inc.

\bibitem[{Buratti et~al.(2020)Buratti, Pujar, Bornea, McCarley, Zheng, Rossiello, Morari, Laredo, Thost, Zhuang, and Domeniconi}]{buratti_exploring_2020}
Luca Buratti, Saurabh Pujar, Mihaela Bornea, Scott McCarley, Yunhui Zheng, Gaetano Rossiello, Alessandro Morari, Jim Laredo, Veronika Thost, Yufan Zhuang, and Giacomo Domeniconi. 2020.
\newblock \href {https://doi.org/10.48550/arXiv.2006.12641} {Exploring {Software} {Naturalness} through {Neural} {Language} {Models}}.
\newblock ArXiv:2006.12641 [cs].

\bibitem[{Campos(2021)}]{campos_curriculum_2021}
Daniel Campos. 2021.
\newblock \href {https://doi.org/10.48550/arXiv.2108.02170} {Curriculum learning for language modeling}.
\newblock ArXiv:2108.02170 [cs].

\bibitem[{Chen et~al.(2021)Chen, Tworek, Jun, Yuan, Pinto, Kaplan, Edwards, Burda, Joseph, Brockman, Ray, Puri, Krueger, Petrov, Khlaaf, Sastry, Mishkin, Chan, Gray, Ryder, Pavlov, Power, Kaiser, Bavarian, Winter, Tillet, Such, Cummings, Plappert, Chantzis, Barnes, Herbert-Voss, Guss, Nichol, Paino, Tezak, Tang, Babuschkin, Balaji, Jain, Saunders, Hesse, Carr, Leike, Achiam, Misra, Morikawa, Radford, Knight, Brundage, Murati, Mayer, Welinder, McGrew, Amodei, McCandlish, Sutskever, and Zaremba}]{chen_evaluating_2021}
Mark Chen, Jerry Tworek, Heewoo Jun, Qiming Yuan, Henrique Ponde de~Oliveira Pinto, Jared Kaplan, Harri Edwards, Yuri Burda, Nicholas Joseph, Greg Brockman, Alex Ray, Raul Puri, Gretchen Krueger, Michael Petrov, Heidy Khlaaf, Girish Sastry, Pamela Mishkin, Brooke Chan, Scott Gray, Nick Ryder, Mikhail Pavlov, Alethea Power, Lukasz Kaiser, Mohammad Bavarian, Clemens Winter, Philippe Tillet, Felipe~Petroski Such, Dave Cummings, Matthias Plappert, Fotios Chantzis, Elizabeth Barnes, Ariel Herbert-Voss, William~Hebgen Guss, Alex Nichol, Alex Paino, Nikolas Tezak, Jie Tang, Igor Babuschkin, Suchir Balaji, Shantanu Jain, William Saunders, Christopher Hesse, Andrew~N. Carr, Jan Leike, Josh Achiam, Vedant Misra, Evan Morikawa, Alec Radford, Matthew Knight, Miles Brundage, Mira Murati, Katie Mayer, Peter Welinder, Bob McGrew, Dario Amodei, Sam McCandlish, Ilya Sutskever, and Wojciech Zaremba. 2021.
\newblock \href {https://doi.org/10.48550/arXiv.2107.03374} {Evaluating {Large} {Language} {Models} {Trained} on {Code}}.
\newblock ArXiv:2107.03374 [cs].

\bibitem[{Deshpande et~al.(2023)Deshpande, Pechi, Thatte, Lialin, and Rumshisky}]{deshpande_honey_2023}
Vijeta Deshpande, Dan Pechi, Shree Thatte, Vladislav Lialin, and Anna Rumshisky. 2023.
\newblock \href {https://doi.org/10.18653/v1/2023.findings-acl.326} {Honey, {I} {Shrunk} the {Language}: {Language} {Model} {Behavior} at {Reduced} {Scale}.}
\newblock In \emph{Findings of the {Association} for {Computational} {Linguistics}: {ACL} 2023}, pages 5298--5314, Toronto, Canada. Association for Computational Linguistics.

\bibitem[{Elman(1993)}]{elman_learning_1993}
Jeffrey~L. Elman. 1993.
\newblock \href {https://doi.org/10.1016/0010-0277(93)90058-4} {Learning and development in neural networks: the importance of starting small}.
\newblock \emph{Cognition}, 48(1):71--99.

\bibitem[{Feng et~al.(2020)Feng, Guo, Tang, Duan, Feng, Gong, Shou, Qin, Liu, Jiang, and Zhou}]{feng_codebert_2020}
Zhangyin Feng, Daya Guo, Duyu Tang, Nan Duan, Xiaocheng Feng, Ming Gong, Linjun Shou, Bing Qin, Ting Liu, Daxin Jiang, and Ming Zhou. 2020.
\newblock \href {https://doi.org/10.18653/v1/2020.findings-emnlp.139} {{CodeBERT}: {A} {Pre}-{Trained} {Model} for {Programming} and {Natural} {Languages}}.
\newblock In \emph{Findings of the {Association} for {Computational} {Linguistics}: {EMNLP} 2020}, pages 1536--1547, Online. Association for Computational Linguistics.

\bibitem[{Guo et~al.(2020)Guo, Ren, Lu, Feng, Tang, Liu, Zhou, Duan, Svyatkovskiy, Fu, Tufano, Deng, Clement, Drain, Sundaresan, Yin, Jiang, and Zhou}]{guo_graphcodebert_2020}
Daya Guo, Shuo Ren, Shuai Lu, Zhangyin Feng, Duyu Tang, Shujie Liu, Long Zhou, Nan Duan, Alexey Svyatkovskiy, Shengyu Fu, Michele Tufano, Shao~Kun Deng, Colin Clement, Dawn Drain, Neel Sundaresan, Jian Yin, Daxin Jiang, and Ming Zhou. 2020.
\newblock \href {https://openreview.net/forum?id=jLoC4ez43PZ} {{GraphCodeBERT}: {Pre}-training {Code} {Representations} with {Data} {Flow}}.

\bibitem[{Halstead(1977)}]{halstead_elements_1977}
Maurice~Howard Halstead. 1977.
\newblock \emph{Elements of software science}.
\newblock Elsevier {Computer} {Science} {Library}. {Operating} and {Programming} {Systems} {Series}. Elsevier, New York.
\newblock OCLC: 908930065.

\bibitem[{Hindle et~al.(2016)Hindle, Barr, Gabel, Su, and Devanbu}]{hindle_naturalness_2016}
Abram Hindle, Earl~T. Barr, Mark Gabel, Zhendong Su, and Premkumar Devanbu. 2016.
\newblock \href {https://doi.org/10.1145/2902362} {On the naturalness of software}.
\newblock \emph{Communications of the ACM}, 59(5):122--131.

\bibitem[{Kanade et~al.(2020)Kanade, Maniatis, Balakrishnan, and Shi}]{kanade_learning_2020}
Aditya Kanade, Petros Maniatis, Gogul Balakrishnan, and Kensen Shi. 2020.
\newblock \href {https://proceedings.mlr.press/v119/kanade20a.html} {Learning and {Evaluating} {Contextual} {Embedding} of {Source} {Code}}.
\newblock In \emph{Proceedings of the 37th {International} {Conference} on {Machine} {Learning}}, pages 5110--5121. PMLR.
\newblock ISSN: 2640-3498.

\bibitem[{Karpathy(2022)}]{karpathy_karpathynanogpt_2022}
Andrej Karpathy. 2022.
\newblock \href {https://github.com/karpathy/nanoGPT} {karpathy/{nanoGPT}}.

\bibitem[{Li et~al.(2022{\natexlab{a}})Li, Zhang, and He}]{li_stability-efficiency_2022}
Conglong Li, Minjia Zhang, and Yuxiong He. 2022{\natexlab{a}}.
\newblock The {Stability}-{Efficiency} {Dilemma}: {Investigating} {Sequence} {Length} {Warmup} for {Training} {GPT} {Models}.
\newblock \emph{Advances in Neural Information Processing Systems}, 35:26736--26750.

\bibitem[{Li et~al.(2022{\natexlab{b}})Li, Choi, Chung, Kushman, Schrittwieser, Leblond, Eccles, Keeling, Gimeno, Lago, Hubert, Choy, d'Autume, Babuschkin, Chen, Huang, Welbl, Gowal, Cherepanov, Molloy, Mankowitz, Robson, Kohli, de~Freitas, Kavukcuoglu, and Vinyals}]{li_competition-level_2022}
Yujia Li, David Choi, Junyoung Chung, Nate Kushman, Julian Schrittwieser, Rémi Leblond, Tom Eccles, James Keeling, Felix Gimeno, Agustin~Dal Lago, Thomas Hubert, Peter Choy, Cyprien de~Masson d'Autume, Igor Babuschkin, Xinyun Chen, Po-Sen Huang, Johannes Welbl, Sven Gowal, Alexey Cherepanov, James Molloy, Daniel~J. Mankowitz, Esme~Sutherland Robson, Pushmeet Kohli, Nando de~Freitas, Koray Kavukcuoglu, and Oriol Vinyals. 2022{\natexlab{b}}.
\newblock \href {https://doi.org/10.1126/science.abq1158} {Competition-{Level} {Code} {Generation} with {AlphaCode}}.
\newblock \emph{Science}, 378(6624):1092--1097.
\newblock ArXiv:2203.07814 [cs].

\bibitem[{Lu et~al.(2021)Lu, Guo, Ren, Huang, Svyatkovskiy, Blanco, Clement, Drain, Jiang, Tang, Li, Zhou, Shou, Zhou, Tufano, Gong, Zhou, Duan, Sundaresan, Deng, Fu, and Liu}]{lu_codexglue_2021}
Shuai Lu, Daya Guo, Shuo Ren, Junjie Huang, Alexey Svyatkovskiy, Ambrosio Blanco, Colin Clement, Dawn Drain, Daxin Jiang, Duyu Tang, Ge~Li, Lidong Zhou, Linjun Shou, Long Zhou, Michele Tufano, Ming Gong, Ming Zhou, Nan Duan, Neel Sundaresan, Shao~Kun Deng, Shengyu Fu, and Shujie Liu. 2021.
\newblock \href {https://datasets-benchmarks-proceedings.neurips.cc/paper/2021/hash/c16a5320fa475530d9583c34fd356ef5-Abstract-round1.html} {{CodeXGLUE}: {A} {Machine} {Learning} {Benchmark} {Dataset} for {Code} {Understanding} and {Generation}}.
\newblock \emph{Proceedings of the Neural Information Processing Systems Track on Datasets and Benchmarks}, 1.

\bibitem[{McCabe(1976)}]{mccabe_complexity_1976}
T.J. McCabe. 1976.
\newblock \href {https://doi.org/10.1109/TSE.1976.233837} {A {Complexity} {Measure}}.
\newblock \emph{IEEE Transactions on Software Engineering}, SE-2(4):308--320.
\newblock Conference Name: IEEE Transactions on Software Engineering.

\bibitem[{Nagatsuka et~al.(2021)Nagatsuka, Broni-Bediako, and Atsumi}]{nagatsuka_pre-training_2021}
Koichi Nagatsuka, Clifford Broni-Bediako, and Masayasu Atsumi. 2021.
\newblock \href {https://aclanthology.org/2021.ranlp-1.112} {Pre-training a {BERT} with {Curriculum} {Learning} by {Increasing} {Block}-{Size} of {Input} {Text}}.
\newblock In \emph{Proceedings of the {International} {Conference} on {Recent} {Advances} in {Natural} {Language} {Processing} ({RANLP} 2021)}, pages 989--996, Held Online. INCOMA Ltd.

\bibitem[{Nagatsuka et~al.(2023)Nagatsuka, Broni-Bediako, and Atsumi}]{nagatsuka_length-based_2023}
Koichi Nagatsuka, Clifford Broni-Bediako, and Masayasu Atsumi. 2023.
\newblock \href {https://doi.org/10.1007/s00354-022-00198-8} {Length-{Based} {Curriculum} {Learning} for {Efficient} {Pre}-training of {Language} {Models}}.
\newblock \emph{New Generation Computing}, 41(1):109--134.

\bibitem[{Nijkamp et~al.(2022)Nijkamp, Pang, Hayashi, Tu, Wang, Zhou, Savarese, and Xiong}]{nijkamp_codegen_2022}
Erik Nijkamp, Bo~Pang, Hiroaki Hayashi, Lifu Tu, Huan Wang, Yingbo Zhou, Silvio Savarese, and Caiming Xiong. 2022.
\newblock \href {https://openreview.net/forum?id=iaYcJKpY2B_} {{CodeGen}: {An} {Open} {Large} {Language} {Model} for {Code} with {Multi}-{Turn} {Program} {Synthesis}}.

\bibitem[{Press et~al.(2021)Press, Smith, and Lewis}]{press_shortformer_2021}
Ofir Press, Noah~A. Smith, and Mike Lewis. 2021.
\newblock \href {https://doi.org/10.48550/arXiv.2012.15832} {Shortformer: {Better} {Language} {Modeling} using {Shorter} {Inputs}}.
\newblock ArXiv:2012.15832 [cs].

\bibitem[{Rozière et~al.(2024)Rozière, Gehring, Gloeckle, Sootla, Gat, Tan, Adi, Liu, Sauvestre, Remez, Rapin, Kozhevnikov, Evtimov, Bitton, Bhatt, Ferrer, Grattafiori, Xiong, Défossez, Copet, Azhar, Touvron, Martin, Usunier, Scialom, and Synnaeve}]{roziere_code_2024}
Baptiste Rozière, Jonas Gehring, Fabian Gloeckle, Sten Sootla, Itai Gat, Xiaoqing~Ellen Tan, Yossi Adi, Jingyu Liu, Romain Sauvestre, Tal Remez, Jérémy Rapin, Artyom Kozhevnikov, Ivan Evtimov, Joanna Bitton, Manish Bhatt, Cristian~Canton Ferrer, Aaron Grattafiori, Wenhan Xiong, Alexandre Défossez, Jade Copet, Faisal Azhar, Hugo Touvron, Louis Martin, Nicolas Usunier, Thomas Scialom, and Gabriel Synnaeve. 2024.
\newblock \href {https://doi.org/10.48550/arXiv.2308.12950} {Code {Llama}: {Open} {Foundation} {Models} for {Code}}.
\newblock ArXiv:2308.12950 [cs].

\bibitem[{Sanger(1994)}]{sanger_neural_1994}
T.D. Sanger. 1994.
\newblock \href {https://doi.org/10.1109/70.294207} {Neural network learning control of robot manipulators using gradually increasing task difficulty}.
\newblock \emph{IEEE Transactions on Robotics and Automation}, 10(3):323--333.
\newblock Conference Name: IEEE Transactions on Robotics and Automation.

\bibitem[{Wang et~al.(2023)Wang, Le, Gotmare, Bui, Li, and Hoi}]{wang_codet5_2023}
Yue Wang, Hung Le, Akhilesh Gotmare, Nghi Bui, Junnan Li, and Steven Hoi. 2023.
\newblock \href {https://doi.org/10.18653/v1/2023.emnlp-main.68} {{CodeT5}+: {Open} {Code} {Large} {Language} {Models} for {Code} {Understanding} and {Generation}}.
\newblock In \emph{Proceedings of the 2023 {Conference} on {Empirical} {Methods} in {Natural} {Language} {Processing}}, pages 1069--1088, Singapore. Association for Computational Linguistics.

\bibitem[{Wang et~al.(2021)Wang, Wang, Joty, and Hoi}]{wang_codet5_2021}
Yue Wang, Weishi Wang, Shafiq Joty, and Steven~C.H. Hoi. 2021.
\newblock \href {https://doi.org/10.18653/v1/2021.emnlp-main.685} {{CodeT5}: {Identifier}-aware {Unified} {Pre}-trained {Encoder}-{Decoder} {Models} for {Code} {Understanding} and {Generation}}.
\newblock In \emph{Proceedings of the 2021 {Conference} on {Empirical} {Methods} in {Natural} {Language} {Processing}}, pages 8696--8708, Online and Punta Cana, Dominican Republic. Association for Computational Linguistics.

\bibitem[{Xu et~al.(2022)Xu, Alon, Neubig, and Hellendoorn}]{xu_systematic_2022}
Frank~F. Xu, Uri Alon, Graham Neubig, and Vincent~Josua Hellendoorn. 2022.
\newblock \href {https://doi.org/10.1145/3520312.3534862} {A systematic evaluation of large language models of code}.
\newblock In \emph{Proceedings of the 6th {ACM} {SIGPLAN} {International} {Symposium} on {Machine} {Programming}}, {MAPS} 2022, pages 1--10, New York, NY, USA. Association for Computing Machinery.

\bibitem[{Xu and Zhu(2022)}]{xu_survey_2022}
Yichen Xu and Yanqiao Zhu. 2022.
\newblock \href {https://doi.org/10.48550/arXiv.2212.10079} {A {Survey} on {Pretrained} {Language} {Models} for {Neural} {Code} {Intelligence}}.
\newblock ArXiv:2212.10079 [cs].

\bibitem[{Yamani et~al.(2024)Yamani, Naïr, and Baghdadi}]{yamani_automatic_2024}
Kamel Yamani, Marwa Naïr, and Riyadh Baghdadi. 2024.
\newblock \href {https://doi.org/10.48550/arXiv.2403.06503} {Automatic {Generation} of {Python} {Programs} {Using} {Context}-{Free} {Grammars}}.
\newblock ArXiv:2403.06503 [cs].

\bibitem[{Zhang et~al.(2021)Zhang, Wei, Wang, Jin, and Cao}]{zhang_reducing_2021}
Wei Zhang, Wei Wei, Wen Wang, Lingling Jin, and Zheng Cao. 2021.
\newblock \href {https://doi.org/10.1109/ISPASS51385.2021.00025} {Reducing {BERT} {Computation} by {Padding} {Removal} and {Curriculum} {Learning}}.
\newblock In \emph{2021 {IEEE} {International} {Symposium} on {Performance} {Analysis} of {Systems} and {Software} ({ISPASS})}, pages 90--92.

\bibitem[{Zheng et~al.(2023)Zheng, Xia, Zou, Dong, Wang, Xue, Wang, Shen, Wang, Li, Su, Yang, and Tang}]{zheng_codegeex_2023}
Qinkai Zheng, Xiao Xia, Xu~Zou, Yuxiao Dong, Shan Wang, Yufei Xue, Zihan Wang, Lei Shen, Andi Wang, Yang Li, Teng Su, Zhilin Yang, and Jie Tang. 2023.
\newblock \href {https://doi.org/10.48550/arXiv.2303.17568} {{CodeGeeX}: {A} {Pre}-{Trained} {Model} for {Code} {Generation} with {Multilingual} {Evaluations} on {HumanEval}-{X}}.
\newblock ArXiv:2303.17568 [cs].

\end{thebibliography}

\appendix

\section{Details about Difficulty Metrics}
\label{sec:difficulty}

\subsection{Cyclomatic Complexity}
Cyclomatic Complexity (CC) \citep{mccabe_complexity_1976} is a software metric used to quantify the number of linearly independent paths through a program's source code. This metric is derived from the program's control flow graph, where nodes symbolize command groups, and directed edges connect nodes if the subsequent command can be executed immediately after the preceding one.

CC is calculated as the number of decisions within a code block, plus one. Specifically, given the control flow graph of a program, the CC metric is calculated using the following formula:

\[ \text{CC} = E - N + 2P \]
where:

\begin{itemize}
    \item \( E \) is the number of edges in the control flow graph.
    \item \( N \) is the number of nodes in the control flow graph.
    \item \( P \) is the number of connected components in the graph (often equal to 1 for a single program).
\end{itemize}

\subsection{Halstead Difficulty}

Halstead's metrics \cite{halstead_elements_1977} aim to quantify various aspects of software, which are computed statically from the source code. In our context, we are particularly interested in the Halstead's Difficulty metric (HD). The following variables are defined:

\begin{itemize}
\item $\eta_1=$ the number of distinct operators
\item $\eta_2=$ the number of distinct operands
\item $N_1=$ the total number of operators
\item $N_2=$ the total number of operands
\end{itemize}

With these variables, we can compute several measures:

\begin{itemize}
\item Program vocabulary: $\eta=\eta_1+\eta_2$
\item Program length: $N=N_1+N_2$
\item Calculated program length: $\widehat{N}=\eta_1 \log _2 \eta_1+\eta_2 \log _2 \eta_2$
\item Volume: $V=N \log _2 \eta$
\item Difficulty: $HD=\frac{\eta_1}{2} \cdot \frac{N_2}{\eta_2}$
\item Effort: $E=D \cdot V$
\item Time required to program: $T=\frac{E}{18} $ seconds
\item Number of delivered bugs: $B=\frac{V}{3000}$
\end{itemize}

\section{Additional Examples of Code Snippets}
\label{sec:examples}

\begin{figure*}[h]
    \centering
    \includegraphics[scale=0.9]{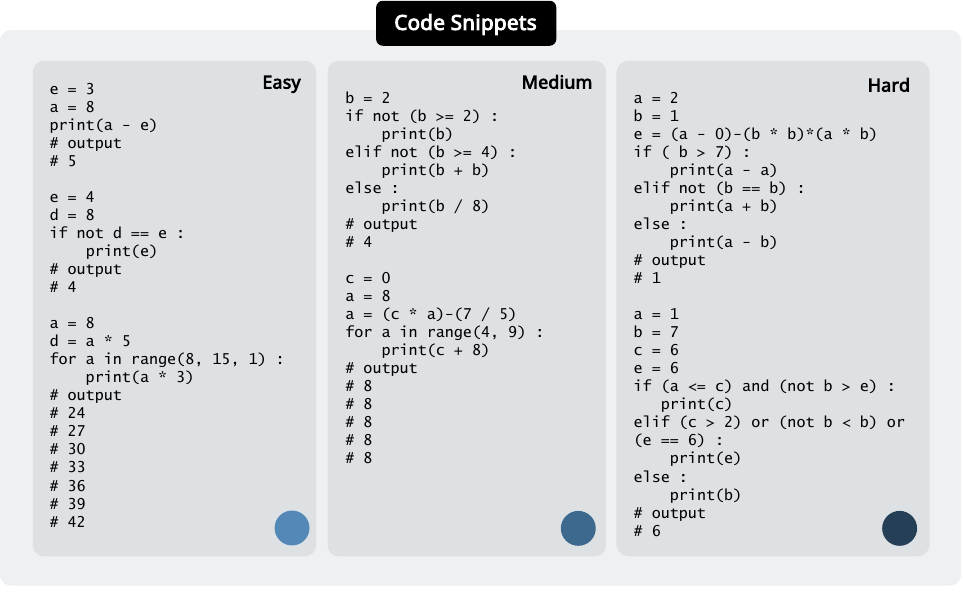}
     \caption{Additional Examples of Code Snippets,}
    \label{fig:add}
\end{figure*}

Additional examples of code snippets are provided in Figure \ref{fig:add}.

\section{Hardware and Software Specifications}
\label{sec:model}
All our models were trained for less than 2 hours on a machine equipped with a single NVIDIA Tesla V100-PCIE-32GB GPU and were implemented using PyTorch 2.0.0. All codes were written in Python 3.8.6.

\section{Generation Process of our models}
\label{sec:generation}
The generation process of our models begins by providing the model with the context, which consists of the last 256 tokens. The model then predicts the logits for the next token based on this context. These logits are converted into a probability distribution via softmax. The \verb|torch.multinomial| function is used to sample the next token from this distribution. This sampled token is added back to the context. This procedure is repeated until the maximum number of new tokens has been generated. The final output consists of all the tokens generated by the model.

\end{document}